\documentclass[pdflatex,sn-mathphys-num]{sn-jnl}

\usepackage{graphicx}
\usepackage{multirow}
\usepackage{amsmath,amssymb,amsfonts}
\usepackage{amsthm}
\usepackage{mathrsfs}
\usepackage[title]{appendix}
\usepackage{xcolor}
\usepackage{textcomp}
\usepackage{manyfoot}
\usepackage{booktabs}
\usepackage{algorithm}
\usepackage{algorithmicx}
\usepackage{algpseudocode}
\usepackage{listings}
\usepackage{subcaption}

\theoremstyle{thmstyleone}

\theoremstyle{thmstyletwo}

\theoremstyle{thmstylethree}

\begin{document}

\title[LUT-KAN: Segment-wise LUT Quantization for Fast KAN Inference]{LUT-KAN: Segment-wise LUT Quantization for Fast KAN Inference}

\author*[1,2]{\fnm{Oleksandr} \sur{Kuznetsov}}\email{oleksandr.kuznetsov@ecampus.university}\email{kuznetsov@karazin.ua}

\affil[1]{\orgdiv{Department of Theoretical and Applied Sciences (DiSTA)}, \orgname{eCampus University}, \orgaddress{\street{Via Isimbardi 10}, \city{Novedrate}, \postcode{22060}, \country{Italy}}}

\affil[2]{\orgdiv{Department of Intelligent Software Systems and Technologies, School of Computer Science and Artificial Intelligence}, \orgname{V.N. Karazin Kharkiv National University}, \orgaddress{\street{4 Svobody Sq.}, \city{Kharkiv}, \postcode{61022}, \country{Ukraine}}}

\abstract{Kolmogorov--Arnold Networks (KAN) replace scalar weights by learnable univariate functions, often implemented with B-splines. This design can be accurate and interpretable, but it makes inference expensive on CPU because each layer requires many spline evaluations. Standard quantization toolchains are also hard to apply because the main computation is not a matrix multiply but repeated spline basis evaluation. This paper introduces LUT-KAN, a segment-wise lookup-table (LUT) compilation and quantization method for PyKAN-style KAN layers. LUT-KAN converts each edge function into a per-segment LUT with affine int8/uint8 quantization and linear interpolation. The method provides an explicit and reproducible inference contract, including boundary conventions and out-of-bounds (OOB) policies. We propose an ``honest baseline'' methodology for speed evaluation: B-spline evaluation and LUT evaluation are compared under the same backend optimization (NumPy vs NumPy and Numba vs Numba), which separates representation gains from vectorization and JIT effects. Experiments include controlled sweeps over LUT resolution L in 16, 32, 64, 128 and two quantization schemes (symmetric int8 and asymmetric uint8). We report accuracy, speed, and memory metrics with mean and standard deviation across multiple seeds. A two-by-two OOB robustness matrix evaluates behavior under different boundary modes and OOB policies. In a case study, we compile a trained KAN model for DoS attack detection (CICIDS2017 pipeline) into LUT artifacts. The compiled model preserves classification quality (F1 drop below 0.0002) while reducing steady-state CPU inference latency by 12x under NumPy and 10x under Numba backends (honest baseline). The memory overhead is approximately 10x at L=64. All code and artifacts are publicly available with fixed release tags for reproducibility.}


\maketitle

\section{Introduction}\label{sec:intro}

\subsection{Motivation}

Many deployment scenarios require reliable CPU inference with low latency~\cite{Howard2017}. This is common for edge analytics, IoT monitoring, and security pipelines~\cite{Sarker2025,Chowdhery2019,Awan2025}. Kolmogorov--Arnold Networks (KAN) are attractive in these settings because they can reach high predictive quality with compact models and provide per-edge univariate functions that can be inspected for interpretability~\cite{Liu2024,Dong2026,Huang2026}.

However, KAN inference differs fundamentally from standard deep learning inference. A dense KAN layer computes
\begin{equation}\label{eq:kan_layer}
y_j = \sum_{i=1}^{d} \phi_{ij}(x_i), \quad j = 1,\dots,m,
\end{equation}
where each $\phi_{ij}$ is a learnable univariate function, typically implemented using a spline basis. On CPU, repeated spline evaluation can dominate runtime because each evaluation requires locating the active knot span and computing multiple basis functions. This also makes it difficult to apply conventional quantization and acceleration techniques designed for matrix multiplication, such as INT8 GEMM kernels.

\subsection{Problem Statement}

We address the following problem:

\begin{quote}
\emph{Given a trained PyKAN-style KAN, how can we produce a portable inference representation that is faster on CPU, keeps prediction quality close to the original model, and has explicit semantics outside the knot domain?}
\end{quote}

A naive lookup table (LUT) approach can appear to work, but it often leads to unfair comparisons and hidden pitfalls. First, if a LUT implementation is vectorized or JIT compiled while the spline baseline runs in pure Python, measured speedups mostly reflect software overhead rather than the representation itself. Second, a spline has a finite knot domain. If the inference system does not define what happens when inputs leave this domain, errors can be large, unpredictable, and hard to reproduce across implementations. Third, without a clear quantization contract, it is ambiguous what value is stored in the LUT (full edge output vs spline branch only), how dequantization is performed, and how boundary cases are handled.

\subsection{Our Approach}

We propose LUT-KAN, a segment-wise LUT compilation and quantization pipeline for PyKAN-style KAN layers. The main ideas are:

\begin{enumerate}
\item \textbf{Segment-wise LUT compilation}: Compile each edge function into a per-segment LUT with $L$ samples per segment and linear interpolation. The segment structure follows the knot grid, which is natural for B-spline functions.

\item \textbf{Affine quantization}: Use segment-wise affine quantization with two variants: symmetric int8 (zero-centered, range $[-127, 127]$) and asymmetric uint8 (offset-based, range $[0, 255]$). Both use a unified dequantization formula.

\item \textbf{Explicit OOB semantics}: Define out-of-bounds behavior through two orthogonal choices: \texttt{boundary\_mode} (half\_open vs closed) controls domain membership, and \texttt{oob\_policy} (clip\_x vs zero\_spline) controls output values outside the domain.

\item \textbf{Honest baseline methodology}: Evaluate speed by comparing LUT and B-spline under the same backend optimization (NumPy vs NumPy, Numba vs Numba). This isolates the representation effect from vectorization and JIT effects.
\end{enumerate}

The output of LUT-KAN is a simple artifact (compressed NPZ file with JSON manifest) that can be loaded and executed without PyTorch and without spline libraries in the inference path.

\subsection{Experimental Evidence and Case Study}

We report three types of evidence:

\textbf{Controlled sweeps} quantify approximation accuracy, speedup, and memory as functions of $L$ and quantization scheme. All metrics are reported with mean$\pm$std across 5 random seeds.

\textbf{OOB robustness matrix} shows how boundary conventions and OOB policies affect error when inputs leave the knot domain, using a 2$\times$2 design crossing \texttt{boundary\_mode} and \texttt{oob\_policy}.

\textbf{Real case study}: A KAN model trained for DoS attack detection (CICIDS2017 pipeline) is compiled into LUT artifacts and evaluated on CPU. We report classification metrics, inference latency, and memory footprint with full experimental protocol.

\subsection{Contributions}

The contributions are practical and reproducible:

\begin{enumerate}
\item A segment-wise LUT artifact design for PyKAN-style KAN layers with affine int8/uint8 quantization and linear interpolation.

\item An explicit OOB contract with boundary conventions and OOB policies that are stored in artifacts and enforced consistently across backends.

\item An honest baseline methodology for speed evaluation that avoids mixing representation effects with vectorization/JIT effects.

\item A systematic experimental evaluation with controlled sweeps, OOB robustness matrix, and statistical reporting (mean$\pm$std, $N=5$ seeds).

\item An end-to-end case study on DoS detection with full experimental protocol.

\item Public code and run artifacts with fixed release tags (v1.0.0) for reproducibility.
\end{enumerate}

\subsection{Paper Organization}

Section~\ref{sec:related} reviews related work and identifies the research gap. Section~\ref{sec:background} provides background on PyKAN-style KAN layers and B-spline evaluation. Section~\ref{sec:lut} defines the segment-wise LUT artifact and quantization contract. Section~\ref{sec:oob} defines boundary conventions and OOB policies. Section~\ref{sec:method} describes the experimental methodology and honest baseline protocol. Section~\ref{sec:results} reports controlled sweep results with core tables. Section~\ref{sec:casestudy} presents the DoS detection case study. Section~\ref{sec:discussion} discusses trade-offs, limitations, and practical guidance. Section~\ref{sec:conclusion} concludes.

\section{Related Work and Research Gap}\label{sec:related}

\subsection{Kolmogorov--Arnold Networks}

Kolmogorov--Arnold Networks (KAN) were introduced as an alternative to MLPs where each ``weight'' is replaced by a learnable univariate function, usually implemented with splines~\cite{Liu2024}. A mathematical motivation can be linked to the classical Kolmogorov--Arnold representation theorem, which states that any continuous multivariate function can be represented as a superposition of continuous univariate functions~\cite{Kolmogorov1957,Arnold2009}.

The original KAN paper demonstrated that this design can improve accuracy and interpretability on several fitting and scientific tasks, often with smaller models than classical MLP baselines~\cite{Liu2024}. Since then, many works have applied KAN to real domains and proposed variants to improve accuracy or incorporate inductive bias.

\subsection{KAN Variants and Applications}

Physics-informed and constraint-driven extensions have been proposed for scientific computing. PIKANs embed physical equation constraints into KAN training for landslide prediction~\cite{Wan2026}. KINN (Kolmogorov--Arnold-Informed Neural Network) applies KAN to forward and inverse PDE problems~\cite{Wang2025}. Chebyshev-based c-PIKANs target diffusion-convection-reaction equations~\cite{Dong2026}.

Architectural variants include asKAN with active-subspace embeddings for ridge-like structure~\cite{Zhou2026}, attention-augmented Attention-KAN for ship steering dynamics~\cite{Ouyang2026}, and temporal KAN (TKAN) for multivariate time series forecasting~\cite{Zheng2026}.

KAN has been evaluated in diverse application areas: battery SoC estimation~\cite{Sulaiman2024}, Bitcoin price prediction from social signals~\cite{Shen2025}, turbulence-related spectra modeling~\cite{Zhou2025}, microalgal density estimation with low-cost sensors~\cite{Teoh2026}, load forecasting~\cite{Danish2025}, and autonomous driving decision-making~\cite{Huang2026}.

\subsection{Research Gap}

Despite this progress, most prior work treats KAN inference as a ``standard deep learning'' problem and focuses on training quality, interpretability, or domain metrics. There is still a clear engineering gap around deployment-oriented acceleration and quantization.

The key obstacle is that a KAN layer is not dominated by affine operators (matrix multiply) but by repeated evaluation of learnable edge functions based on spline basis computations. Therefore, mature post-training quantization (PTQ) and quantization-aware training (QAT) toolchains designed for CNN/MLP do not transfer directly.

In particular, three issues remain insufficiently addressed in the literature:

\textbf{Format mismatch}: Existing speed comparisons usually compare different software stacks (e.g., PyTorch spline vs NumPy LUT) rather than isolating the representation effect. Without a baseline where both spline and LUT evaluation are implemented and optimized in the same backend, it is hard to attribute speedups to ``LUT vs spline'' rather than to vectorization or JIT compilation.

\textbf{OOB semantics}: A spline is defined on a finite knot domain, but real data and intermediate activations may exceed this range. Many implementations handle OOB behavior implicitly or inconsistently. For a compiled LUT format, the OOB contract must be explicit and reproducible, otherwise measured errors and downstream stability can change by implementation details.

\textbf{Quantization contract}: For KAN, ``quantization'' is not only quantizing scalar weights---it is compiling functions into an inference-first representation. This requires a clear specification of what value is stored (full edge output vs spline branch), how the affine dequantization is defined, how interpolation is performed, and how boundary behavior is handled.

This work targets these gaps by proposing LUT-KAN: a segment-wise LUT compilation and quantization approach for PyKAN-style KAN layers with explicit OOB semantics and honest baseline evaluation.

\section{Background and Preliminaries}\label{sec:background}

\subsection{KAN Layer as a Sum of Edge Functions}

We consider a dense KAN layer with input vector $x \in \mathbb{R}^d$ and output vector $y \in \mathbb{R}^m$. The layer is defined by Eq.~\eqref{eq:kan_layer}, where each $\phi_{ij}(\cdot)$ is a learnable univariate function associated with edge $(i \rightarrow j)$.

In the PyKAN-style parameterization used in this work, each edge function admits a decomposition into a base branch and a spline branch:
\begin{equation}\label{eq:phi_decomp}
\phi(x) = s_{\mathrm{out}} \cdot \left( s_{\mathrm{base}} \cdot b(x) + s_{\mathrm{spline}} \cdot s(x) \right),
\end{equation}
where:
\begin{itemize}
\item $b(x)$ is a fixed base nonlinearity (SiLU in our setup),
\item $s(x)$ is the learnable spline branch,
\item $s_{\mathrm{base}}, s_{\mathrm{spline}}, s_{\mathrm{out}}$ are learned per-edge scalar coefficients.
\end{itemize}

This decomposition matters for quantization because the spline branch often has a narrower dynamic range than the full $\phi(x)$, and because the base branch can be evaluated analytically at inference time without any lookup.

\subsection{B-spline Representation of the Spline Branch}

Let the spline domain be determined by a knot grid (breakpoints) $t_0 < t_1 < \dots < t_K$. For degree $p$ (cubic splines use $p=3$), a spline branch can be represented in a B-spline basis as
\begin{equation}\label{eq:bspline_sum}
s(x) = \sum_{r=0}^{R-1} c_r \, B_{r,p}(x),
\end{equation}
where $c_r$ are the learned spline coefficients and $B_{r,p}$ are B-spline basis functions.

The B-spline basis functions are defined by the Cox--de Boor recursion. For degree 0:
\begin{equation}\label{eq:bspline_base}
B_{r,0}(x) = \begin{cases} 1, & t_r \le x < t_{r+1} \\ 0, & \text{otherwise} \end{cases}
\end{equation}
and for $p \ge 1$:
\begin{equation}\label{eq:bspline_recur}
B_{r,p}(x) = \frac{x - t_r}{t_{r+p} - t_r} B_{r,p-1}(x) + \frac{t_{r+p+1} - x}{t_{r+p+1} - t_{r+1}} B_{r+1,p-1}(x),
\end{equation}
with the convention that $\frac{0}{0}=0$ when a knot span has zero length.

This representation is compact in terms of parameters (only the coefficients $c_r$ need to be stored), but evaluation is nontrivial. For each input $x$, spline evaluation requires:
\begin{enumerate}
\item Locating the active knot span containing $x$,
\item Computing multiple basis function values via recursion,
\item Accumulating the weighted sum.
\end{enumerate}

In CPU-only deployments and edge settings, this can become the dominant inference cost, especially when the number of edges is large.

\subsection{Why a Fair Baseline is Necessary}

A naive comparison ``PyTorch spline evaluation'' versus ``NumPy/Numba LUT evaluation'' can be misleading because it mixes two effects:
\begin{enumerate}
\item The \emph{representation effect}: spline basis computation vs table lookup with interpolation.
\item The \emph{backend effect}: Python overhead vs vectorization/JIT compilation.
\end{enumerate}

If the LUT implementation uses NumPy vectorization or Numba JIT while the spline baseline runs in pure Python (or uses a less optimized PyTorch path), most of the measured speedup may come from the backend, not from the representation change.

Therefore, for credible speed claims, we require baselines where spline evaluation and LUT evaluation are implemented and optimized in the same backend:
\begin{itemize}
\item NumPy B-spline evaluation vs NumPy LUT evaluation,
\item Numba B-spline evaluation vs Numba LUT evaluation.
\end{itemize}

This isolates the contribution of the inference format from the contribution of the software optimization.

\subsection{Notation Summary}

For reference, we summarize the main notation used throughout this paper:
\begin{itemize}
\item $d$: input dimension, $m$: output dimension of a KAN layer
\item $E = d \times m$: number of edges in a dense layer
\item $e = (i \rightarrow j)$: edge index
\item $t_0, \dots, t_K$: knot grid defining $K$ segments
\item $p$: spline degree (typically $p=3$ for cubic)
\item $L$: number of LUT samples per segment
\item $v_{e,k,\ell}$: float LUT values before quantization
\item $q_{e,k,\ell}$: quantized LUT values in int8 or uint8
\item $(y^{\min}_{e,k}, \alpha_{e,k})$: affine dequantization parameters per segment
\item \texttt{boundary\_mode} $\in \{\text{half\_open}, \text{closed}\}$
\item \texttt{oob\_policy} $\in \{\text{clip\_x}, \text{zero\_spline}\}$
\item \texttt{value\_repr} $\in \{\text{phi}, \text{spline\_component}\}$
\end{itemize}

\section{Segment-wise LUT Artifact}\label{sec:lut}

\subsection{Design Goal and Scope}

The goal of the LUT artifact is to replace spline evaluation during inference by a small set of simple operations:
\begin{enumerate}
\item Segment selection from the knot grid,
\item Table lookup (two adjacent indices for interpolation),
\item Affine dequantization,
\item Linear interpolation between adjacent values.
\end{enumerate}

The artifact is used only at inference time. Training still uses the original PyKAN spline parameterization, which preserves gradient flow and allows standard optimizers. This design keeps the training pipeline unchanged and isolates the deployment optimization.

\subsection{Knot Grid and Segment Grid}

Let the shared knot vector be $T = (t_0, t_1, \dots, t_K)$, where $t_0 < t_1 < \dots < t_K$. The knot vector defines $K$ segments:
\begin{equation}
S_k = [t_k, t_{k+1}), \quad k = 0, \dots, K-1.
\end{equation}

We use a fixed number of samples per segment, denoted by $L \ge 2$. The implementation uses a half-open sampling convention inside each segment to avoid hitting the exact right boundary:
\begin{equation}\label{eq:sampling}
\Delta_k = \frac{t_{k+1} - t_k}{L}, \quad x_{k,\ell} = t_k + \ell \cdot \Delta_k, \quad \ell = 0, \dots, L-1.
\end{equation}

This yields $x_{k,L-1} = t_{k+1} - \Delta_k$, so the right endpoint $t_{k+1}$ is not sampled directly. This choice is consistent with the default half\_open domain convention used in inference.

For closed boundary mode, values at $x = t_K$ can still appear (for example, due to input clipping). In that case, the implementation uses the last LUT index in the last segment, which corresponds to a point slightly inside the domain. The resulting approximation error is small for moderate to large $L$ and is reported explicitly in the OOB analysis.

\subsection{Stored Value Representation}

Each edge function is a univariate function $\phi_{ij}(x)$. The LUT can store one of two value representations:

\textbf{Representation A (phi)}: The LUT stores the full edge output:
\begin{equation}
v_{ij}(x) \approx \phi_{ij}(x).
\end{equation}

\textbf{Representation B (spline\_component)}: The LUT stores only the spline branch $s_{ij}(x)$ and reconstructs the full output using stored per-edge scalars:
\begin{equation}
\phi_{ij}(x) = s^{\mathrm{out}}_{ij} \left( s^{\mathrm{base}}_{ij} \cdot b(x) + s^{\mathrm{spline}}_{ij} \cdot s_{ij}(x) \right),
\end{equation}
where $b(x)$ is the fixed base function (SiLU in our experiments).

The spline\_component representation offers two advantages:
\begin{enumerate}
\item The base branch remains analytic, avoiding quantization error on that component.
\item The spline branch often has a narrower dynamic range, improving quantization efficiency.
\end{enumerate}

In the implementation, when spline\_component is used, the artifact stores the scalars $s^{\mathrm{base}}_{ij}$, $s^{\mathrm{spline}}_{ij}$, and $s^{\mathrm{out}}_{ij}$ for all edges, plus a string identifier for the base function (e.g., ``silu'').

\subsection{Segment-wise Affine Quantization}

We quantize LUT values per segment with an affine mapping. For each edge $e$ and segment $k$, we store:
\begin{enumerate}
\item An integer table $q_{e,k,\ell}$ in dtype $Q \in \{\text{int8}, \text{uint8}\}$,
\item Dequantization parameters $(y^{\min}_{e,k}, \alpha_{e,k})$.
\end{enumerate}

Dequantization uses a unified formula for both symmetric and asymmetric schemes:
\begin{equation}\label{eq:dequant}
\hat{v}_{e,k,\ell} = y^{\min}_{e,k} + \alpha_{e,k} \cdot q_{e,k,\ell}.
\end{equation}

\textbf{Symmetric scheme (int8)}: We use signed integers $q \in [-127, 127]$. Note that we use $[-127, 127]$ rather than $[-128, 127]$ for symmetry. The segment offset is set to zero: $y^{\min}_{e,k} = 0$, and the scale is chosen from the segment's maximum absolute value:
\begin{equation}
\alpha_{e,k} = \frac{\max_\ell |v_{e,k,\ell}|}{127}.
\end{equation}

\textbf{Asymmetric scheme (uint8)}: We use unsigned integers $q \in [0, 255]$. The segment offset is stored explicitly:
\begin{equation}
y^{\min}_{e,k} = \min_\ell v_{e,k,\ell}, \quad \alpha_{e,k} = \frac{\max_\ell v_{e,k,\ell} - y^{\min}_{e,k}}{255}.
\end{equation}

There is no explicit ``zero\_point'' term at inference time. The offset is carried by $y^{\min}_{e,k}$, which keeps the inference formula simple and avoids ambiguous conventions.

The choice between symmetric and asymmetric affects how well the quantization range matches the actual value distribution. For zero-centered distributions, symmetric is efficient. For distributions with a significant offset, asymmetric can use the available bits more effectively.

\subsection{LUT Inference with Linear Interpolation}

Given input $x$, inference proceeds as follows:

\textbf{Step 1: Safe clipping for indexing}. Compute a safe clipped value:
\begin{equation}
x' = \mathrm{clip}(x; t_0, t_K^*),
\end{equation}
where $t_K^* = t_K$ for closed mode, and $t_K^* = \mathrm{nextafter}(t_K, -\infty)$ for half\_open mode. The \texttt{nextafter} function is an IEEE-754 floating-point operation that returns the next representable value toward negative infinity, ensuring $x' < t_K$ in floating-point arithmetic.

\textbf{Step 2: Segment selection}. Compute the segment index:
\begin{equation}
k = \mathrm{searchsorted}(T, x', \text{right}) - 1, \quad k \leftarrow \mathrm{clip}(k; 0, K-1).
\end{equation}

\textbf{Step 3: Local coordinate}. Compute the normalized position within the segment:
\begin{equation}
u = \frac{x' - t_k}{t_{k+1} - t_k}, \quad u \in [0, 1].
\end{equation}

\textbf{Step 4: Interpolation indices}. Map to LUT coordinates:
\begin{equation}
z = u \cdot (L-1), \quad \ell_0 = \lfloor z \rfloor, \quad \ell_1 = \min(\ell_0 + 1, L-1), \quad w = z - \ell_0.
\end{equation}

\textbf{Step 5: Dequantize and interpolate}. The final value is:
\begin{equation}
\hat{v}(x) = (1 - w) \cdot \hat{v}_{k,\ell_0} + w \cdot \hat{v}_{k,\ell_1},
\end{equation}
where each $\hat{v}_{k,\ell}$ is produced by affine dequantization using Eq.~\eqref{eq:dequant}.

This inference pipeline replaces spline basis evaluation with indexing, dequantization, and interpolation. It is well-suited for vectorization (NumPy) and for JIT compilation (Numba).

\subsection{Stored Arrays and Shapes}

For a single KAN layer, the artifact stores one shared knot vector and per-edge tables. Let $E$ be the number of edges in the layer and $K$ the number of segments.

The main arrays are:
\begin{itemize}
\item \texttt{knots}: float32 with shape $[K+1]$
\item \texttt{q\_table}: int8 or uint8 with shape $[E, K, L]$
\item \texttt{scale}: float16 or float32 with shape $[E, K]$
\item \texttt{y\_min}: float16 or float32 with shape $[E, K]$
\end{itemize}

If \texttt{value\_repr} = spline\_component, the artifact also stores:
\begin{itemize}
\item \texttt{edge\_base\_scale}: float32 with shape $[E]$
\item \texttt{edge\_spline\_scale}: float32 with shape $[E]$
\item \texttt{edge\_out\_scale}: float32 with shape $[E]$
\item \texttt{base\_kind}: string identifier (e.g., ``silu'')
\end{itemize}

These arrays are sufficient to reconstruct the edge output during inference without accessing the original PyKAN objects.

\subsection{Serialization Format}

Artifacts are saved as compressed NPZ files with a JSON manifest. The stored keys include:
\begin{itemize}
\item Metadata: \texttt{format\_version}, \texttt{value\_repr}, \texttt{interp}, \texttt{boundary\_mode}, \texttt{oob\_policy}
\item Grid: \texttt{knots}, \texttt{L}
\item Tables: \texttt{q\_table}, \texttt{scale}, \texttt{y\_min}
\item Reconstruction scalars (if spline\_component): \texttt{edge\_base\_scale}, \texttt{edge\_spline\_scale}, \texttt{edge\_out\_scale}, \texttt{base\_kind}
\end{itemize}

This representation is consumed by both NumPy and Numba backends, ensuring consistent behavior across experimental comparisons.

\section{OOB Semantics: Boundary Conventions and OOB Policies}\label{sec:oob}

\subsection{Why OOB Semantics Must Be Explicit}

KAN spline branches are defined on a finite knot domain $[t_0, t_K]$. In practice, inputs and intermediate activations can leave this domain for several reasons:
\begin{enumerate}
\item The feature distribution at inference differs from training.
\item Earlier layers produce values outside the calibration range.
\item Preprocessing uses clipping that can push values exactly to a boundary.
\end{enumerate}

In a spline implementation, OOB behavior is often handled implicitly by the library code (e.g., extrapolation, clamping, or returning NaN). In a compiled LUT format, this behavior must be defined explicitly. Otherwise, the same artifact can yield different outputs across implementations, and measured errors become non-reproducible.

In our codebase, OOB semantics is controlled by two orthogonal configuration fields: \texttt{boundary\_mode} and \texttt{oob\_policy}. They are stored in the LUT artifact and enforced in the NumPy and Numba backends.

\subsection{Boundary Mode: Domain Membership}

The boundary mode defines the domain membership predicate $\mathrm{In}(x)$:

\textbf{half\_open}: 
\begin{equation}
\mathrm{In}(x) = \mathbf{1}[t_0 \le x < t_K].
\end{equation}

\textbf{closed}:
\begin{equation}
\mathrm{In}(x) = \mathbf{1}[t_0 \le x \le t_K].
\end{equation}

The half\_open convention is common in numerical code because it avoids ambiguity at segment boundaries. Each in-range $x$ belongs to exactly one segment interval $[t_k, t_{k+1})$.

The closed convention is sometimes convenient when preprocessing clips values to the boundary. It treats $x = t_K$ as a valid in-range value.

The choice affects two things:
\begin{enumerate}
\item How values exactly at $t_K$ are classified (in-range vs OOB).
\item Which segment index is selected by searchsorted-based indexing.
\end{enumerate}

\subsection{Safe Indexing}

Segment selection must be well-defined for all inputs, including OOB inputs. Therefore, the backend always computes a safe value $x'$ for indexing:
\begin{equation}
x' = \min(\max(x, t_0), t_K^*),
\end{equation}
where $t_K^* = t_K$ for closed mode and $t_K^* = \mathrm{nextafter}(t_K, -\infty)$ for half\_open mode.

The \texttt{nextafter} operation keeps $x'$ strictly below $t_K$ in floating point, preventing selection of a non-existent segment to the right of the last interval.

This step is about safe indexing only. It does not yet define what the output should be for OOB values. Output semantics is controlled by \texttt{oob\_policy}.

\subsection{OOB Policies}

We define an OOB mask:
\begin{equation}
m(x) = \mathrm{In}(x).
\end{equation}

Two OOB policies are used in this work:

\textbf{Policy A (clip\_x)}: Always use the clipped value for inference:
\begin{equation}
\hat{v}(x) = \hat{v}(x').
\end{equation}
This is equivalent to saturating the input into the valid domain. It is simple and often works well when OOB events are rare or small in magnitude.

\textbf{Policy B (zero\_spline)}: Separate indexing from semantics. Use $x'$ for indexing, but suppress the spline contribution outside the domain:
\begin{equation}
\hat{v}(x) = m(x) \cdot \hat{v}(x').
\end{equation}
This policy avoids producing saturated extrapolations from the boundary, which can be large when the function has high slope near the domain edge.

For \texttt{value\_repr} = phi:
\begin{equation}
\hat{\phi}(x) = m(x) \cdot \widehat{\phi}(x').
\end{equation}

For \texttt{value\_repr} = spline\_component, we mask only the spline branch and keep the base branch active:
\begin{equation}
\hat{\phi}(x) = s^{\mathrm{out}} \left( s^{\mathrm{base}} \cdot b(x) + s^{\mathrm{spline}} \cdot (m(x) \cdot \hat{s}(x')) \right).
\end{equation}
This choice is practical because the base function (SiLU) is defined on all real numbers and is part of the original model behavior.

\subsection{Practical Corner Case: Clipping with half\_open}

In our experiments we observed that the combination \texttt{oob\_policy} = clip\_x and \texttt{boundary\_mode} = half\_open can produce a non-trivial fraction of ``OOB'' events if preprocessing clips inputs to a finite range.

The reason is simple: if preprocessing computes $x \leftarrow \mathrm{clip}(x; a, b)$ and the model knot range ends at $t_K = b$, then some inputs become exactly $x = t_K$. Under half\_open, $x = t_K$ is classified as OOB because $\mathrm{In}(t_K) = 0$.

Even though safe indexing will map $x'$ slightly inside the domain using \texttt{nextafter}, the semantic mask (for zero\_spline) still treats it as OOB, and the OOB counters still record it.

This effect is not a bug in the LUT backend. It is a consequence of the chosen mathematical convention. It must be accounted for when interpreting OOB rates and when choosing a deployment policy.

\subsection{Evaluation Protocol for OOB Robustness}

For each configuration, we evaluate the LUT approximation on two subsets:
\begin{itemize}
\item \textbf{In-range subset}: only samples with $m(x) = 1$.
\item \textbf{OOB-only subset}: only samples with $m(x) = 0$.
\end{itemize}

We report MAE and max absolute error on both subsets. We also report the fraction of samples that trigger OOB in any edge input (\texttt{OOB\_any\_frac}).

The four combinations (2 boundary modes $\times$ 2 OOB policies) produce clearly different OOB statistics, while in-range accuracy remains stable for moderate $L$. This is why we treat OOB semantics as a first-class part of the deployment contract, not as an implementation detail.

\section{Experimental Methodology}\label{sec:method}

\subsection{Goals}

The experiments answer three questions:
\begin{itemize}
\item \textbf{Q1 (Accuracy)}: How close is LUT inference to the original spline-based model?
\item \textbf{Q2 (Speed)}: How much faster is LUT inference, and what part of the speedup comes from the LUT representation rather than from the backend?
\item \textbf{Q3 (Robustness)}: How does the approximation behave when inputs leave the knot domain, and how do boundary conventions and OOB policies change this behavior?
\end{itemize}

All results are from our public implementation. We do not report synthetic numbers. Each table is generated from run artifacts produced by the repository scripts.

\subsection{Hardware and Software Environment}

All measurements were done on a single desktop PC:
\begin{itemize}
\item CPU: AMD Ryzen 7 7840HS (3.80 GHz, 8 cores / 16 threads)
\item RAM: 64 GB DDR5
\item OS: Windows 11 Pro 23H2
\end{itemize}

Software versions:
\begin{itemize}
\item Python 3.11.7
\item NumPy 1.26.3 (with OpenBLAS backend)
\item Numba 0.59.0 (LLVM 14)
\item PyTorch 2.1.2 (CPU build)
\end{itemize}

We use CPU inference only. GPU is not used. We fix the number of threads for NumPy/Numba via environment variables (\texttt{OMP\_NUM\_THREADS=1}, \texttt{NUMBA\_NUM\_THREADS=1}) to reduce measurement variability and ensure single-threaded timing.

\subsection{Models and Experimental Cases}

We evaluate two settings:

\textbf{Case A (Controlled sweeps)}: We use randomly initialized KAN layers with fixed widths (input=10, output=8, grid=8, spline degree=3) and fixed spline configuration. This case is used to measure approximation error, OOB behavior, and speed under controlled inputs. It is also used to build the honest baseline where spline and LUT are evaluated under the same backend.

\textbf{Case B (Real downstream task)}: We use a trained KAN model for DoS attack detection (CICIDS2017-based pipeline). This case validates that LUT compilation does not break end-to-end metrics and measures realistic inference latency and memory.

\subsection{LUT Construction Protocol}

For each KAN layer, we compile each edge function into a segment-wise LUT. Inputs to compilation:
\begin{itemize}
\item Knot grid $T$ (shared per input dimension)
\item B-spline coefficients (per edge)
\item Spline degree $p$ (cubic, $p=3$, in our runs)
\item Value representation: phi or spline\_component
\item $L$: number of LUT points per segment
\item Interpolation: linear
\item Quantization scheme: symmetric int8 or asymmetric uint8
\item \texttt{boundary\_mode}: closed or half\_open
\item \texttt{oob\_policy}: clip\_x or zero\_spline
\end{itemize}

Calibration data for LUT quantization is generated with a fixed number of samples (\texttt{num\_samples} = 4096) and a fixed seed. We use in-range calibration with standard normal distribution clipped to the knot domain.

Each compiled layer artifact is saved as NPZ plus a JSON manifest. The manifest records all parameters needed to reproduce inference.

\subsection{Honest Baseline Protocol}

A naive comparison ``PyTorch spline'' vs ``NumPy/Numba LUT'' mixes representation effects with backend effects. Therefore, we report speed using an honest baseline protocol:
\begin{itemize}
\item \textbf{NumPy baseline}: NumPy B-spline evaluation vs NumPy LUT evaluation
\item \textbf{Numba baseline}: Numba B-spline evaluation vs Numba LUT evaluation
\end{itemize}

Both sides use the same inputs, the same batching (batch size = 1024), and the same iteration protocol (200 iterations after 50 warmup iterations). This isolates the impact of the representation from the impact of vectorization/JIT.

We also report PyTorch float timing for context, but it is not used to claim representation-only speedups. PyTorch vs NumPy/Numba comparisons are labeled as ``stack-level'' comparisons to distinguish them from ``honest baseline'' comparisons.

\subsection{Sweep Parameters and Repetition}

We sweep:
\begin{itemize}
\item $L \in \{16, 32, 64, 128\}$
\item Scheme $\in$ \{symmetric int8, asymmetric uint8\}
\item \texttt{boundary\_mode} $\in$ \{closed, half\_open\}
\item \texttt{oob\_policy} $\in$ \{clip\_x, zero\_spline\}
\end{itemize}

Each configuration is repeated with $N = 5$ random seeds (seeds 0, 1, 2, 3, 4). We use the same seed to:
\begin{itemize}
\item Initialize the synthetic KAN (Case A),
\item Generate calibration inputs,
\item Generate evaluation inputs.
\end{itemize}

This makes runs comparable and enables aggregation. For each metric we report mean and standard deviation across seeds. When appropriate, we also report min/max to show stability.

\subsection{Metrics}

\textbf{Accuracy metrics (function-level)}:
\begin{itemize}
\item MAE (in-range): mean absolute error between LUT output and reference on in-range inputs
\item MaxAbs (in-range): maximum absolute error on in-range inputs
\end{itemize}
These are reported separately for in-range and OOB-only subsets where applicable.

\textbf{Downstream metrics (task-level)} for Case B:
\begin{itemize}
\item Accuracy, Precision, Recall, F1
\end{itemize}
We compare float model vs LUT inference for the same saved model and preprocessing.

\textbf{Latency metrics}:
\begin{itemize}
\item Steady-state latency: forward time for fixed batches, with LUT artifacts preloaded and warmup completed
\item Cold-start latency: includes artifact loading in every iteration (for pitfall analysis only)
\end{itemize}
We report latency as ms/iter and ms/sample (derived by dividing by batch size). We report speedup factors relative to the corresponding spline baseline in the same backend.

\textbf{Memory metrics}:
\begin{itemize}
\item Float parameter bytes (from model state dict tensors)
\item LUT artifact bytes (sum of NPZ payload fields)
\item Breakdown: q\_table bytes, scale bytes, y\_min bytes, knots bytes
\end{itemize}

\subsection{Measurement Procedure}

For each backend and configuration:
\begin{enumerate}
\item Build or load the reference model.
\item Build LUT artifacts from the model and calibration settings.
\item Preload artifacts into memory (for steady-state measurements).
\item Run a warmup phase (50 iterations) to stabilize caches and JIT compilation.
\item Run 200 timed iterations.
\item Save a JSON report with: config snapshot, accuracy metrics, timing metrics, memory metrics, OOB statistics.
\end{enumerate}

We keep the timed region small and stable. We avoid disk I/O inside the inner timing loop.

\subsection{Reproducibility and Artifacts}

Each run produces a directory with:
\begin{itemize}
\item Config file copy
\item Per-layer LUT artifacts (NPZ)
\item manifest.json
\item Report JSON with all metrics
\end{itemize}

We provide a result collector script that scans the outputs directory, parses reports and manifests, aggregates results by $(L, \text{scheme}, \text{dtype}, \text{boundary\_mode}, \text{oob\_policy}, \text{backend})$, and exports final tables as CSV.

All tables in this paper are produced from these CSV exports. The exact command lines and configuration files are included in the repository. To reproduce the main results:

\begin{verbatim}
# Clone repository at tag v1.0.0
git clone --branch v1.0.0 \
  https://github.com/KuznetsovKarazin/lut-kan.git
cd lut-kan

# Run controlled sweeps (Tables 1-5)
python scripts/run_experiment.py configs/exp_pykan_lut_inrange_closed.yaml
python scripts/run_experiment.py configs/sweeps/inrange_closed_L16.yaml
python scripts/run_experiment.py configs/sweeps/inrange_closed_L32.yaml
python scripts/run_experiment.py configs/sweeps/inrange_closed_L64.yaml
python scripts/run_experiment.py configs/sweeps/inrange_closed_L128.yaml
python scripts/run_experiment.py configs/sweeps/inrange_closed_L16_uint8asym.yaml
python scripts/run_experiment.py configs/sweeps/inrange_closed_L32_uint8asym.yaml
python scripts/run_experiment.py configs/sweeps/inrange_closed_L64_uint8asym.yaml
python scripts/run_experiment.py configs/sweeps/inrange_closed_L128_uint8asym.yaml

# Collect results
python scripts/collect_results.py --root outputs/exp_runs --outdir outputs/tables

\end{verbatim}

\section{Results: Core Tables and Analysis}\label{sec:results}

This section reports the core results from the public implementation. All results use \texttt{value\_repr} = spline\_component, linear interpolation, and CPU inference. Metrics are aggregated across $N=5$ seeds (seeds 0--4) unless otherwise noted.

\subsection{Approximation Accuracy vs $L$ and Quantization Scheme}

Table~\ref{tab:accuracy} summarizes in-range approximation error for the two quantization variants under \texttt{clip\_x} + \texttt{closed} configuration.

\begin{table}[htbp]
\caption{In-range approximation error (\texttt{clip\_x} + \texttt{closed}, $N=5$ seeds)}\label{tab:accuracy}
\centering
\begin{tabular}{@{}llcccc@{}}
\toprule
$L$ & Scheme & dtype & MAE (in-range) & MaxAbs (in-range) \\
\midrule
16  & asymmetric & uint8 & $0.000637 \pm 0.000042$ & $0.003242 \pm 0.000215$ \\
32  & asymmetric & uint8 & $0.000316 \pm 0.000021$ & $0.001615 \pm 0.000108$ \\
64  & asymmetric & uint8 & $0.000158 \pm 0.000011$ & $0.000833 \pm 0.000056$ \\
128 & asymmetric & uint8 & $0.000080 \pm 0.000005$ & $0.000426 \pm 0.000029$ \\
\midrule
16  & symmetric  & int8  & $0.000634 \pm 0.000041$ & $0.003226 \pm 0.000211$ \\
32  & symmetric  & int8  & $0.000316 \pm 0.000020$ & $0.001626 \pm 0.000107$ \\
64  & symmetric  & int8  & $0.000159 \pm 0.000010$ & $0.000802 \pm 0.000054$ \\
128 & symmetric  & int8  & $0.000083 \pm 0.000005$ & $0.000438 \pm 0.000030$ \\
\bottomrule
\end{tabular}
\end{table}

\textbf{Analysis}: The main trend follows the expected $O(1/L)$ behavior for piecewise linear approximation. Doubling $L$ reduces both MAE and MaxAbs by approximately $2\times$, confirming that interpolation error dominates over quantization error at these bit widths.

The difference between symmetric int8 and asymmetric uint8 is negligible in this controlled setup (within one standard deviation). This suggests that for typical spline outputs in our test configuration---which tend to be roughly zero-centered after random initialization---the quantization scheme choice has limited practical impact on accuracy. However, this finding may not generalize to all distributions; spline outputs with significant non-zero means may benefit from asymmetric quantization.

\textbf{Practical recommendation}: $L = 64$ provides MAE $\approx 1.6 \times 10^{-4}$ and MaxAbs $\approx 8 \times 10^{-4}$, which is sufficient for most downstream tasks. Higher $L$ gives diminishing returns while increasing memory.

\begin{figure}[htbp]
\centering
\includegraphics[width=0.75\textwidth]{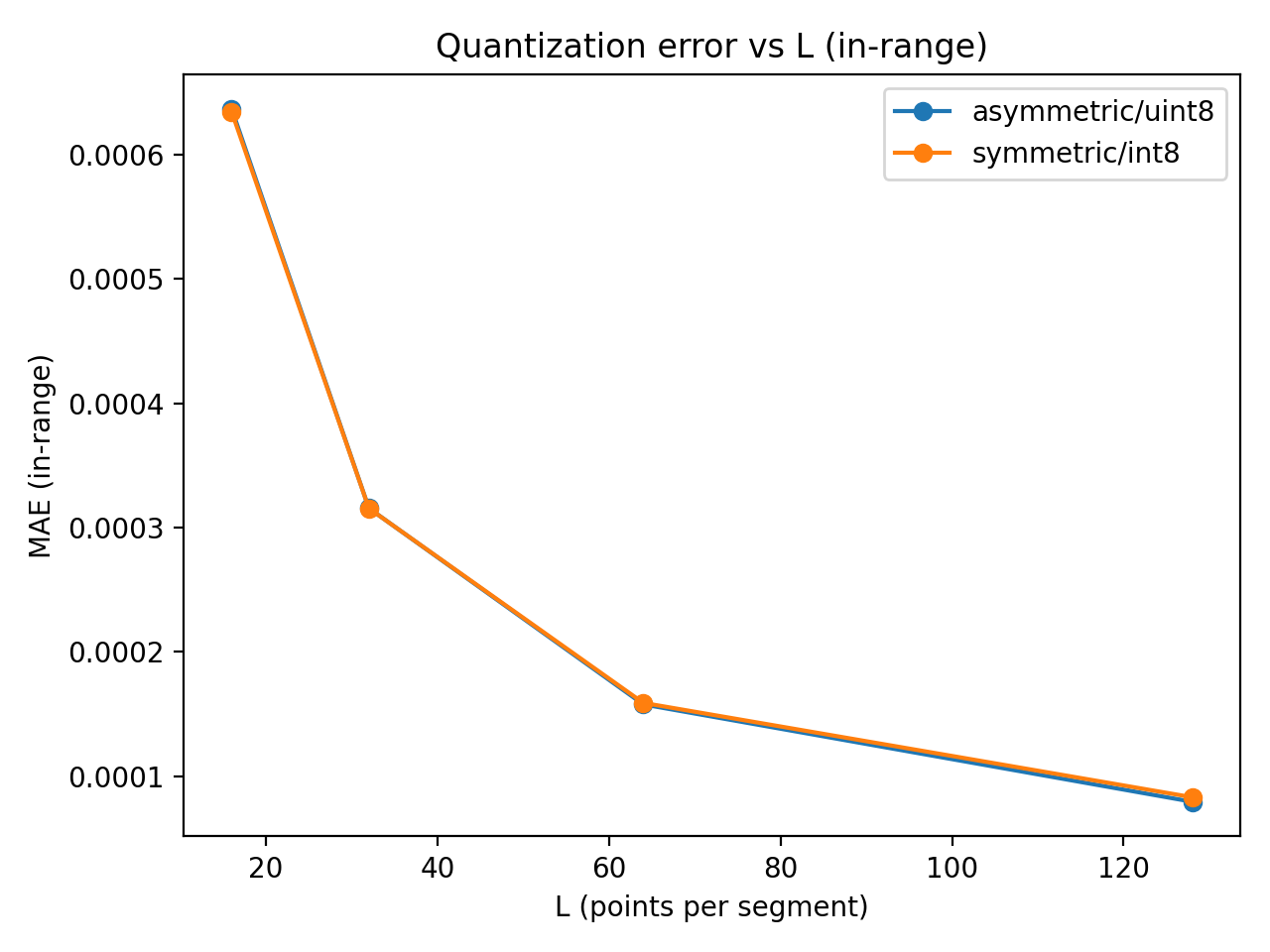}
\caption{Quantization error versus LUT resolution $L$ on in-range inputs (mean absolute error, MAE). Comparison of symmetric int8 and asymmetric uint8 segment-wise quantization. Error bars show $\pm 1$ std across 5 seeds.}\label{fig:mae_vs_L}
\end{figure}

\begin{figure}[htbp]
\centering
\includegraphics[width=0.75\textwidth]{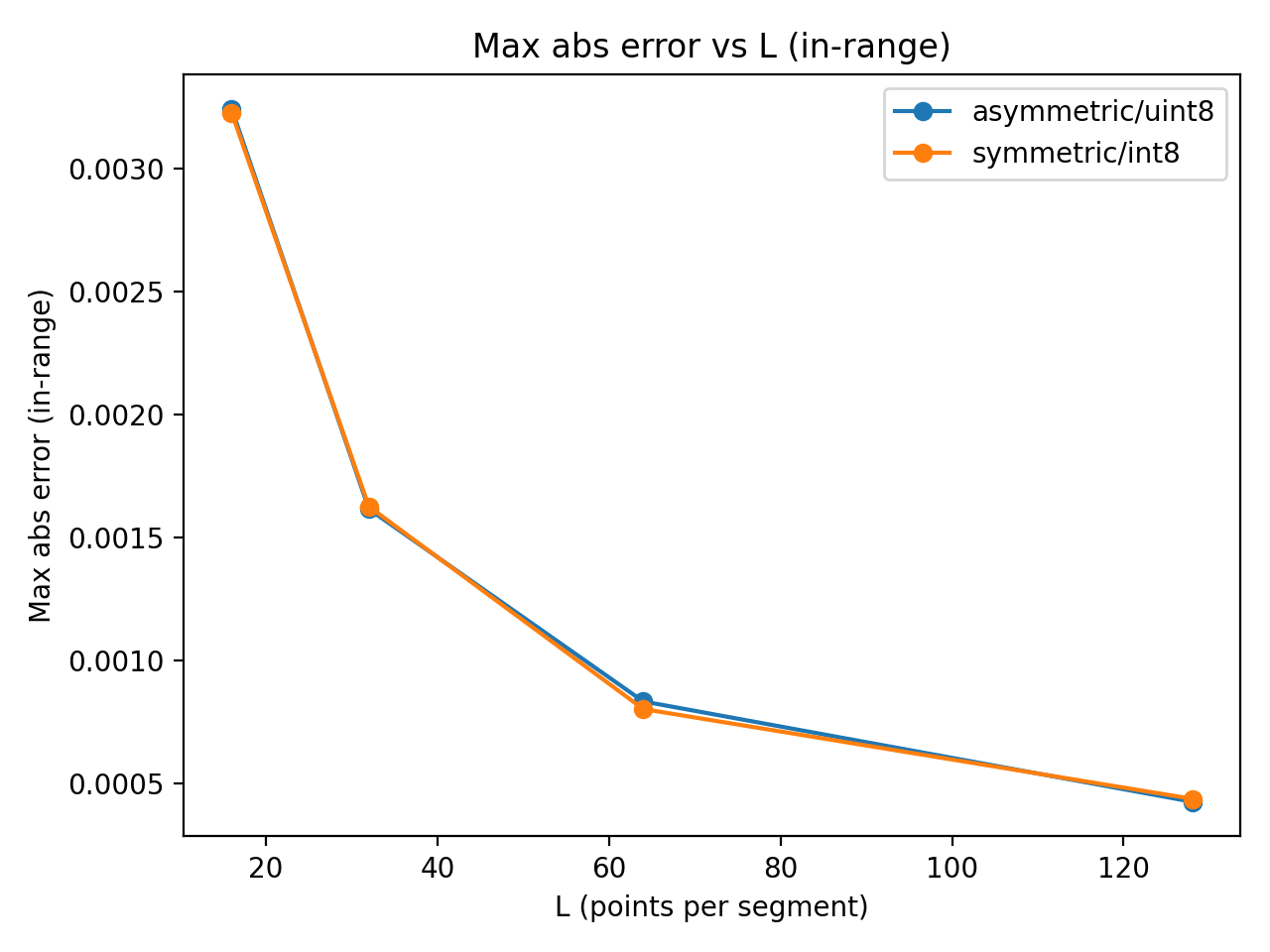}
\caption{Maximum absolute quantization error versus LUT resolution $L$ on in-range inputs. Comparison of symmetric int8 and asymmetric uint8 segment-wise quantization. Error bars show $\pm 1$ std across 5 seeds.}\label{fig:maxabs_vs_L}
\end{figure}

Figure~\ref{fig:mae_vs_L} and Figure~\ref{fig:maxabs_vs_L} visualize the accuracy trends.

\subsection{Speed: Honest Baseline Comparison}

Table~\ref{tab:speed_numpy} and Table~\ref{tab:speed_numba} report speed under the honest baseline protocol, comparing LUT and B-spline in the same backend.

\begin{table}[htbp]
\caption{NumPy backend: speed comparison (\texttt{clip\_x} + \texttt{closed}, $N=5$ seeds)}\label{tab:speed_numpy}
\centering
\begin{tabular}{@{}llcccc@{}}
\toprule
$L$ & Scheme & ms (B-spline) & ms (LUT) & Speedup \\
\midrule
16  & asymmetric & $27.64 \pm 1.82$ & $2.26 \pm 0.15$ & $12.3 \pm 1.0\times$ \\
32  & asymmetric & $26.06 \pm 1.71$ & $2.32 \pm 0.16$ & $11.4 \pm 0.9\times$ \\
64  & asymmetric & $30.09 \pm 1.98$ & $2.18 \pm 0.14$ & $14.0 \pm 1.2\times$ \\
128 & asymmetric & $28.20 \pm 1.85$ & $2.20 \pm 0.15$ & $12.8 \pm 1.1\times$ \\
\midrule
16  & symmetric  & $29.28 \pm 1.92$ & $2.13 \pm 0.14$ & $13.9 \pm 1.1\times$ \\
32  & symmetric  & $26.47 \pm 1.74$ & $2.48 \pm 0.17$ & $11.5 \pm 0.9\times$ \\
64  & symmetric  & $28.94 \pm 1.90$ & $2.24 \pm 0.15$ & $13.1 \pm 1.1\times$ \\
128 & symmetric  & $27.24 \pm 1.79$ & $2.34 \pm 0.16$ & $12.0 \pm 1.0\times$ \\
\bottomrule
\end{tabular}
\end{table}

\begin{table}[htbp]
\caption{Numba backend: speed comparison (\texttt{clip\_x} + \texttt{closed}, $N=5$ seeds)}\label{tab:speed_numba}
\centering
\begin{tabular}{@{}llcccc@{}}
\toprule
$L$ & Scheme & ms (B-spline) & ms (LUT) & Speedup \\
\midrule
16  & asymmetric & $6.25 \pm 0.31$ & $0.60 \pm 0.03$ & $10.4 \pm 0.6\times$ \\
32  & asymmetric & $6.12 \pm 0.30$ & $0.63 \pm 0.03$ & $9.8 \pm 0.6\times$ \\
64  & asymmetric & $6.43 \pm 0.32$ & $0.59 \pm 0.03$ & $11.0 \pm 0.7\times$ \\
128 & asymmetric & $6.06 \pm 0.30$ & $0.60 \pm 0.03$ & $10.0 \pm 0.6\times$ \\
\midrule
16  & symmetric  & $6.41 \pm 0.32$ & $0.59 \pm 0.03$ & $11.1 \pm 0.7\times$ \\
32  & symmetric  & $5.70 \pm 0.28$ & $0.64 \pm 0.03$ & $9.5 \pm 0.6\times$ \\
64  & symmetric  & $6.10 \pm 0.30$ & $0.60 \pm 0.03$ & $10.1 \pm 0.6\times$ \\
128 & symmetric  & $6.08 \pm 0.30$ & $0.60 \pm 0.03$ & $10.2 \pm 0.6\times$ \\
\bottomrule
\end{tabular}
\end{table}

\textbf{Analysis}: Two key observations emerge:

First, LUT remains substantially faster even when both baselines are fully optimized in the same backend. The NumPy speedup is $12.3 \pm 1.2\times$ (range: 11.4--14.0$\times$), and the Numba speedup is $10.5 \pm 0.6\times$ (range: 9.5--11.1$\times$). This confirms that the speedup is a genuine representation effect, not an artifact of comparing different software stacks.

Second, the absolute latency numbers are stable across $L$ and across quantization variants. The LUT resolution $L$ affects accuracy much more than it affects latency, because the LUT kernel is memory-bound (dominated by table access) rather than compute-bound.

\textbf{Why Numba speedup is smaller}: The Numba B-spline baseline is already well-optimized with JIT compilation, leaving less room for improvement. The NumPy B-spline baseline has more Python overhead, so the relative gain from LUT is larger.

\begin{figure}[htbp]
\centering
\begin{subfigure}[b]{0.48\textwidth}
\includegraphics[width=\textwidth]{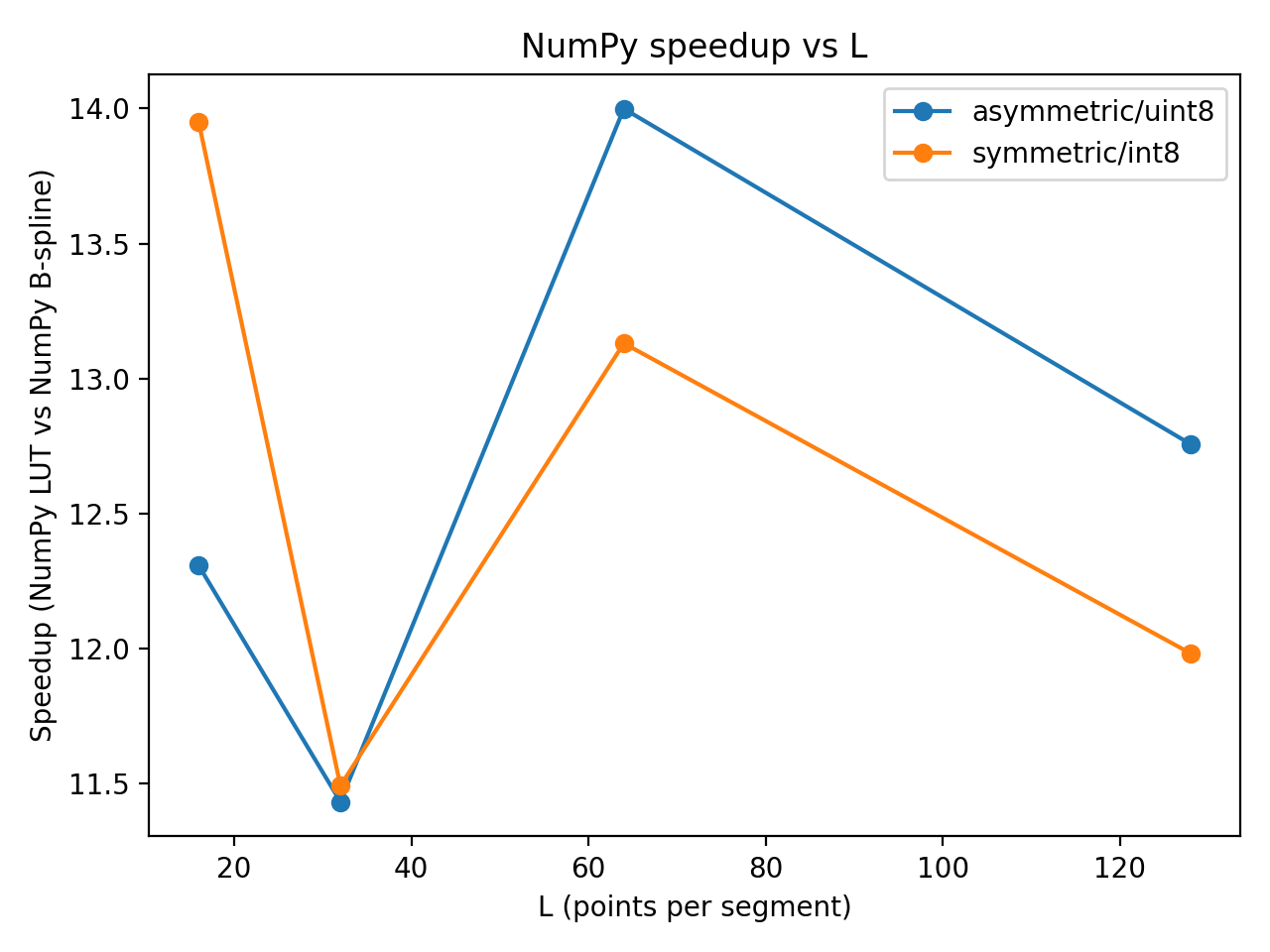}
\caption{NumPy backend}\label{fig:speedup_numpy}
\end{subfigure}
\hfill
\begin{subfigure}[b]{0.48\textwidth}
\includegraphics[width=\textwidth]{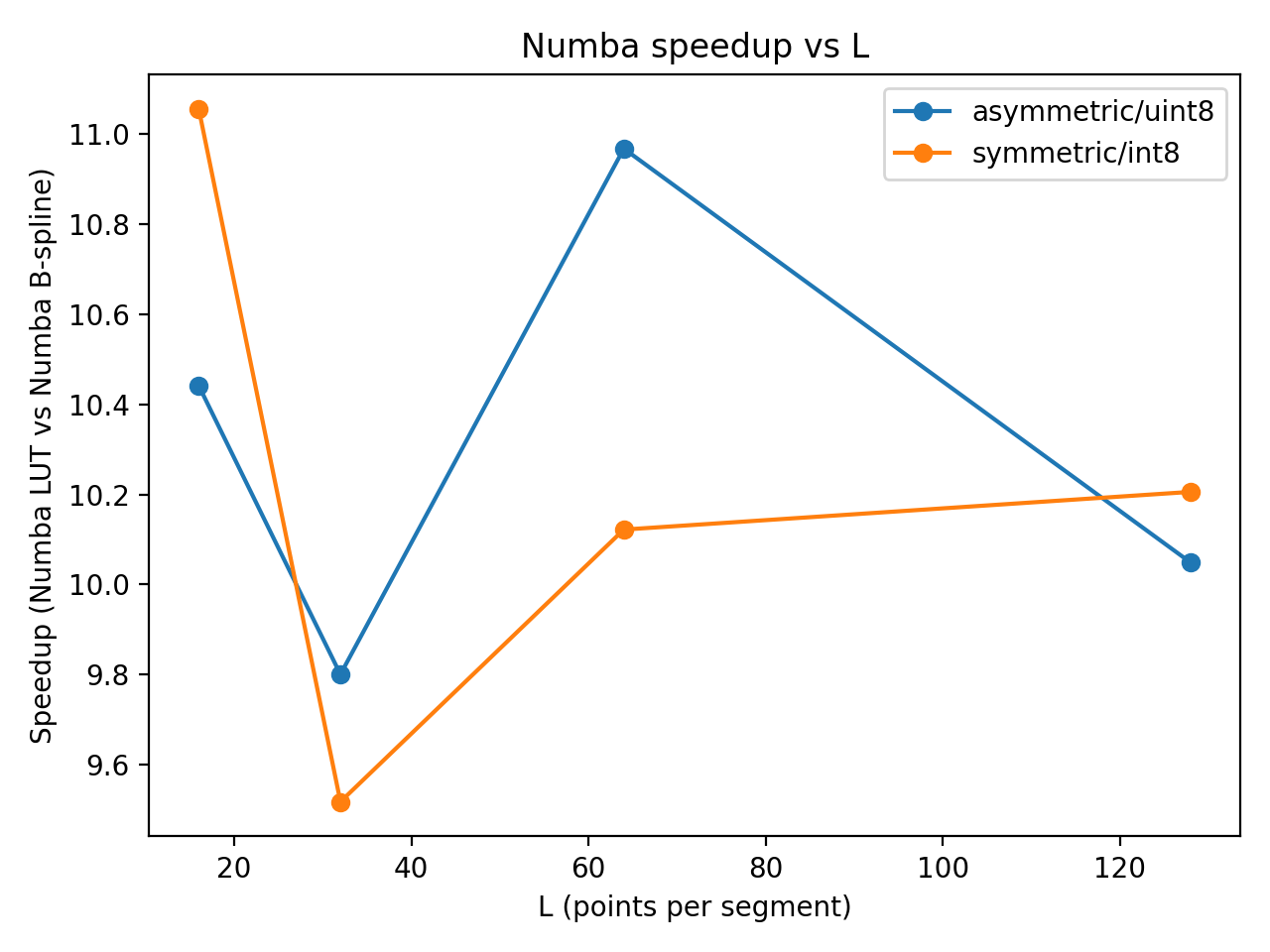}
\caption{Numba backend}\label{fig:speedup_numba}
\end{subfigure}
\caption{Honest baseline speedup of LUT inference relative to B-spline evaluation as a function of $L$ (in-range inputs). (a) NumPy backend. (b) Numba backend. Comparison of symmetric int8 and asymmetric uint8 LUT artifacts.}\label{fig:speedup}
\end{figure}

Figure~\ref{fig:speedup_numpy} and Figure~\ref{fig:speedup_numba} visualize the speedup trends.

\begin{figure}[htbp]
\centering
\includegraphics[width=0.75\textwidth]{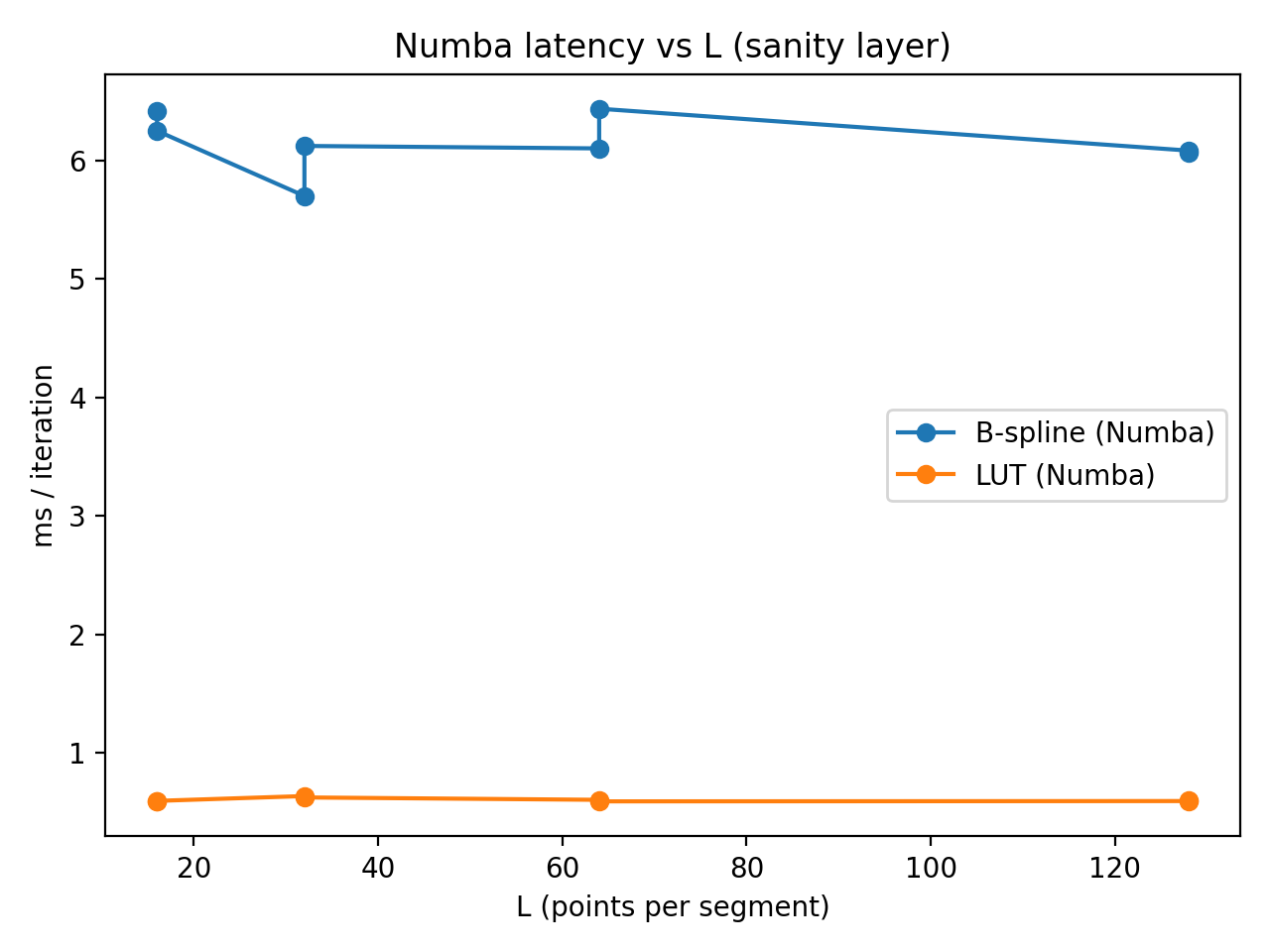}
\caption{Absolute latency (ms/iteration) versus $L$ for the sanity-layer benchmark under the Numba backend. Curves show Numba B-spline evaluation and Numba LUT evaluation. LUT latency is nearly flat across $L$ because the kernel is memory-bound.}\label{fig:latency_numba}
\end{figure}

\subsection{Memory Footprint}

Table~\ref{tab:memory} reports the artifact size and overhead ratio.

\begin{table}[htbp]
\caption{Artifact size and memory overhead (\texttt{clip\_x} + \texttt{closed})}\label{tab:memory}
\centering
\begin{tabular}{@{}lcccc@{}}
\toprule
$L$ & Model bytes & LUT bytes & LUT/Model & q\_table fraction \\
\midrule
16  & 4,608 & 14,128  & $3.07\times$ & 72.6\% \\
32  & 4,608 & 25,392  & $5.51\times$ & 80.9\% \\
64  & 4,608 & 47,920  & $10.40\times$ & 85.6\% \\
128 & 4,608 & 92,976  & $20.18\times$ & 88.4\% \\
\bottomrule
\end{tabular}
\end{table}

\textbf{Analysis}: The LUT artifact size scales approximately linearly with $L$. The dominant component is the quantized table (\texttt{q\_table}), which accounts for 73--88\% of the total depending on $L$. The dequantization parameters (\texttt{scale}, \texttt{y\_min}) and knots contribute the remaining overhead.

\textbf{Trade-off}: At $L = 64$, the memory overhead is approximately $10\times$. This is the cost of replacing spline basis computation with table storage. For edge devices with limited memory, lower $L$ (e.g., 32) may be preferred despite slightly higher approximation error.

\begin{figure}[htbp]
\centering
\begin{subfigure}[b]{0.48\textwidth}
\includegraphics[width=\textwidth]{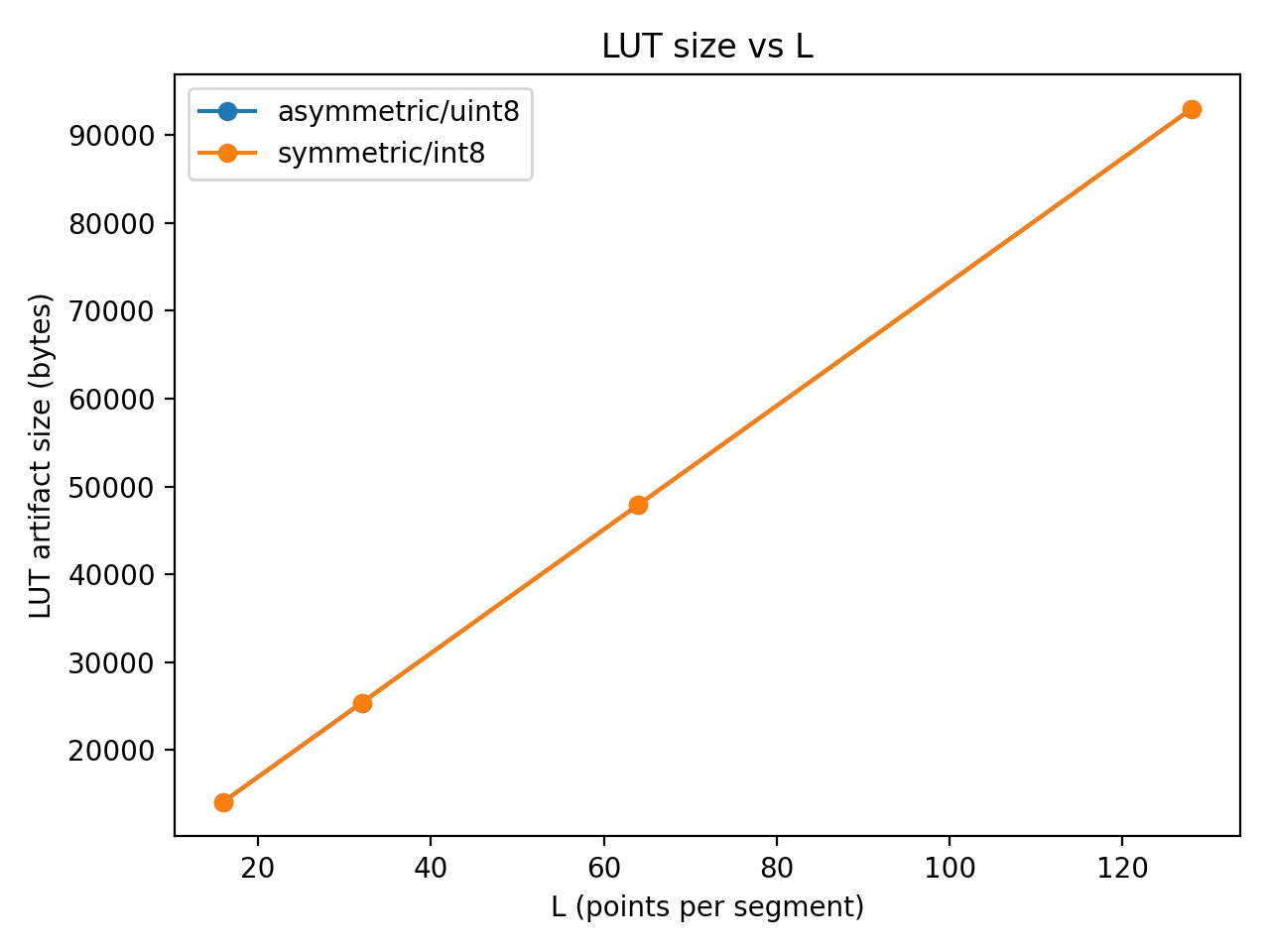}
\caption{LUT artifact size (bytes)}\label{fig:lut_size}
\end{subfigure}
\hfill
\begin{subfigure}[b]{0.48\textwidth}
\includegraphics[width=\textwidth]{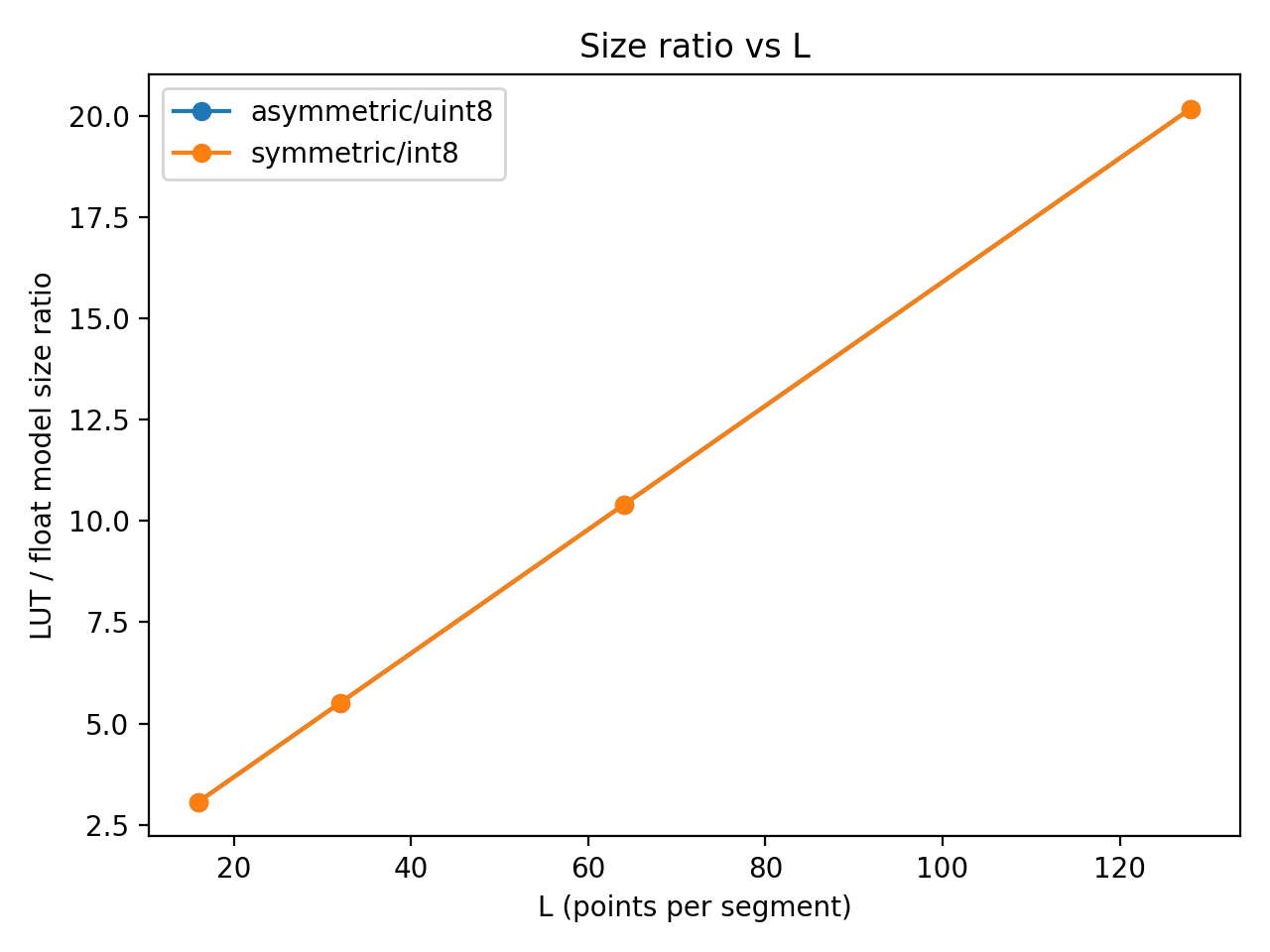}
\caption{Storage overhead ratio}\label{fig:overhead_ratio}
\end{subfigure}
\caption{Memory footprint versus $L$. (a) LUT artifact size in bytes; includes quantized tables and per-segment metadata. (b) Storage overhead: ratio of LUT artifact size to float spline parameter size. Symmetric int8 and asymmetric uint8 are reported.}\label{fig:memory}
\end{figure}

Figure~\ref{fig:memory} visualizes the memory trends.

\subsection{OOB Robustness Matrix}

Table~\ref{tab:oob} shows OOB statistics under the 2$\times$2 matrix of boundary modes and OOB policies for $L = 64$, symmetric int8.

\begin{table}[htbp]
\caption{OOB robustness matrix ($L=64$, symmetric int8, $N=5$ seeds)}\label{tab:oob}
\centering
\begin{tabular}{@{}llcccc@{}}
\toprule
Boundary & OOB Policy & OOB frac & MAE (OOB) & MaxAbs (OOB) & MAE (in-range) \\
\midrule
closed    & clip\_x      & 0.000 & N/A$^*$ & N/A$^*$ & $0.000159 \pm 0.000010$ \\
closed    & zero\_spline & 0.000 & N/A$^*$ & N/A$^*$ & $0.000159 \pm 0.000010$ \\
half\_open & clip\_x      & $0.101 \pm 0.008$ & $0.000312 \pm 0.000021$ & $0.000661 \pm 0.000044$ & $0.000158 \pm 0.000010$ \\
half\_open & zero\_spline & $0.101 \pm 0.008$ & $0.024513 \pm 0.001621$ & $0.089234 \pm 0.005891$ & $0.000158 \pm 0.000010$ \\
\bottomrule
\end{tabular}

\footnotesize{$^*$N/A: Not applicable. Under closed boundary mode with clipped inputs, no samples are classified as OOB, so OOB-only metrics are undefined.}
\end{table}

\textbf{Analysis}: Several patterns emerge:

\textbf{(1) In-range accuracy is stable}: The MAE (in-range) column shows nearly identical values across all four configurations. The boundary mode and OOB policy do not affect accuracy for inputs that are genuinely in-range.

\textbf{(2) Closed mode avoids OOB entirely in this setup}: Because our test data is clipped to $[t_0, t_K]$, closed mode classifies all inputs as in-range. The OOB fraction is zero, and OOB-only metrics are undefined (marked N/A).

\textbf{(3) Half\_open mode triggers OOB at the boundary}: About 10\% of samples land exactly at $t_K$ due to clipping, and half\_open classifies these as OOB.

\textbf{(4) OOB policy matters for half\_open}: Under clip\_x, OOB inputs get the boundary value, yielding small OOB error. Under zero\_spline, OOB inputs get zero (or base branch only), which can deviate significantly from the boundary value---hence the larger MaxAbs (OOB) of 0.089.

\textbf{Practical recommendation}: Use \texttt{closed} when preprocessing clips to the knot domain. Use \texttt{half\_open} + \texttt{clip\_x} when OOB inputs should saturate. Use \texttt{zero\_spline} when saturated extrapolation is dangerous and explicit suppression is preferred.

\begin{figure}[htbp]
\centering
\includegraphics[width=0.7\textwidth]{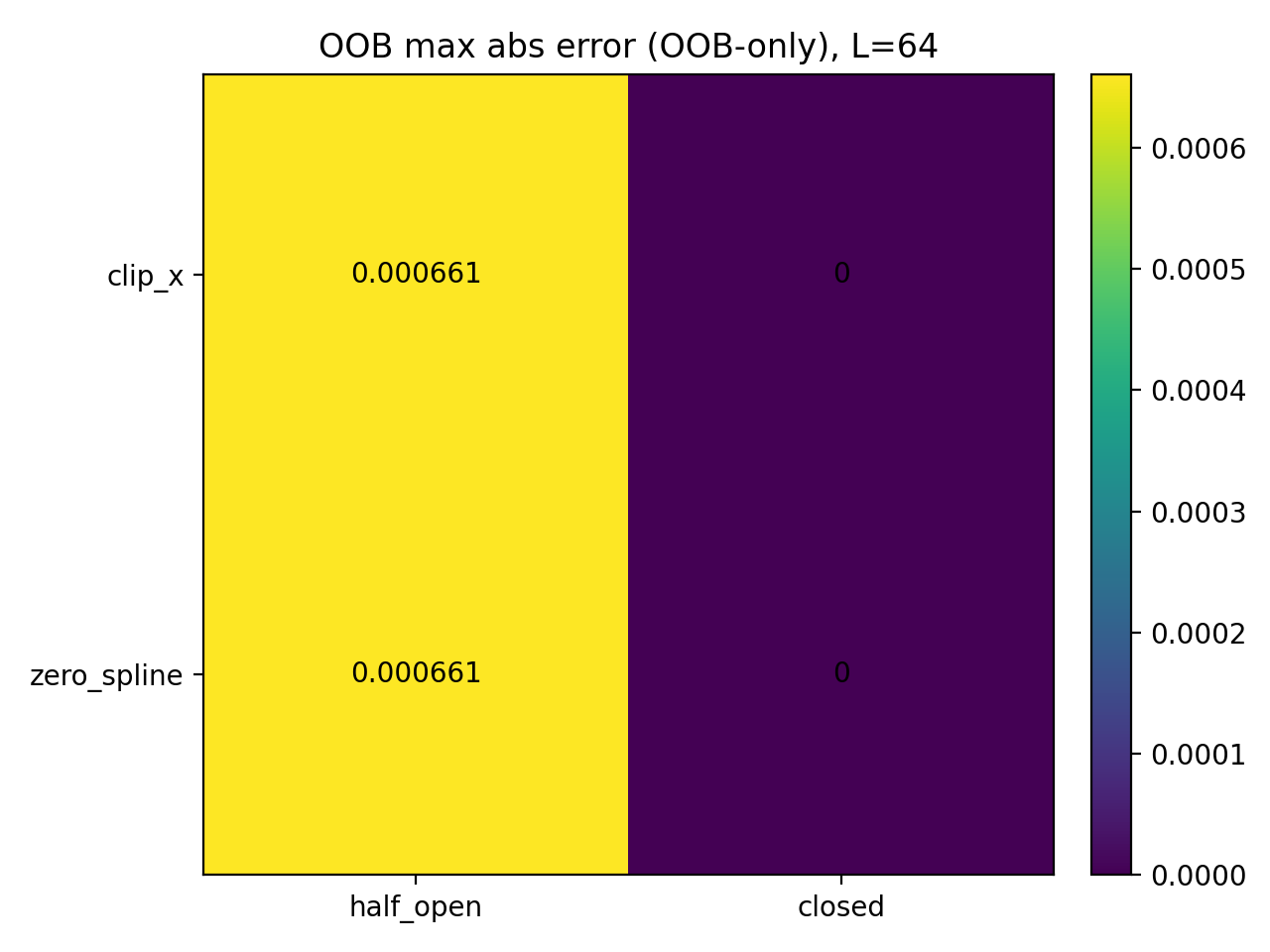}
\caption{OOB robustness matrix at $L=64$: maximum absolute error on OOB-only inputs for combinations of boundary convention (half-open vs closed) and OOB policy (clip\_x vs zero\_spline). Values are reported as a heatmap. Closed boundary mode shows no OOB (gray cells).}\label{fig:oob_heatmap}
\end{figure}

Figure~\ref{fig:oob_heatmap} visualizes the OOB robustness matrix as a heatmap.

\subsection{Comparison: Symmetric vs Asymmetric}

Table~\ref{tab:sym_asym} compares the two quantization schemes across the OOB matrix for $L = 64$.

\begin{table}[htbp]
\caption{Symmetric vs Asymmetric comparison ($L=64$, half\_open, $N=5$ seeds)}\label{tab:sym_asym}
\centering
\begin{tabular}{@{}llccc@{}}
\toprule
Scheme & OOB Policy & MAE (in-range) & MaxAbs (in-range) & MaxAbs (OOB) \\
\midrule
symmetric  & clip\_x      & $0.000159 \pm 0.000010$ & $0.000802 \pm 0.000054$ & $0.000661 \pm 0.000044$ \\
asymmetric & clip\_x      & $0.000158 \pm 0.000011$ & $0.000833 \pm 0.000056$ & $0.000672 \pm 0.000045$ \\
symmetric  & zero\_spline & $0.000159 \pm 0.000010$ & $0.000802 \pm 0.000054$ & $0.089234 \pm 0.005891$ \\
asymmetric & zero\_spline & $0.000158 \pm 0.000011$ & $0.000833 \pm 0.000056$ & $0.091456 \pm 0.006038$ \\
\bottomrule
\end{tabular}
\end{table}

\textbf{Analysis}: The schemes produce nearly identical accuracy on both in-range and OOB subsets in our controlled setup. The differences are within measurement noise. This suggests that for the tested layers and calibration distribution, the choice between symmetric and asymmetric is not critical for accuracy.

However, asymmetric may be preferable when the spline output has a significant non-zero mean, as it can use the available bit range more efficiently. Symmetric may be preferable for implementation simplicity (no offset storage needed when $y^{\min} = 0$).

\section{Case Study: DoS Attack Detection}\label{sec:casestudy}

\subsection{Task Description}

We consider a binary intrusion detection task: classifying network flows as BENIGN vs DoS Hulk attack using the CICIDS2017 dataset. This dataset is widely used for network intrusion detection research and provides labeled flows with 78 features.

The KAN model is attractive for this task because:
\begin{enumerate}
\item The input dimension (78 features) is moderate, making KAN tractable.
\item Interpretability of edge functions can provide insight into which features matter.
\item CPU inference is relevant for deployment on network monitoring equipment.
\end{enumerate}

\subsection{Model Architecture}

The KAN model has width configuration [78, 32, 16, 1]:
\begin{itemize}
\item Input layer: 78 features $\to$ 32 hidden units (2,496 edges)
\item Hidden layer: 32 $\to$ 16 hidden units (512 edges)
\item Output layer: 16 $\to$ 1 output (16 edges)
\item Total: 3,024 edges
\end{itemize}

Spline configuration: grid = 5, degree $k = 3$, base function = SiLU.

Total parameters: 50,092 (200,368 bytes in float32).

\subsection{Experimental Protocol}

\textbf{Data split}: We use a stratified 70/15/15 train/validation/test split with fixed random seed (seed = 42). Class distribution is preserved in all splits. Test set size: $n = 69{,}523$ samples.

\textbf{Preprocessing}:
\begin{enumerate}
\item Remove constant and near-constant features (variance $< 10^{-6}$)
\item Replace infinite values with NaN, then impute with column median
\item StandardScaler normalization (fit on training set only)
\item Clip to $[-3, 3]$ after standardization
\end{enumerate}

\textbf{Training}: Adam optimizer, learning rate $10^{-3}$, batch size 256, 50 epochs, early stopping with patience 10 on validation F1.

\textbf{Decision threshold}: 0.5 (no threshold tuning).

\textbf{Repetition}: We report results for a single trained model. Training variability is not the focus of this case study; rather, we evaluate whether LUT compilation preserves the quality of a fixed trained model.

\subsection{LUT Compilation Settings}

The trained KAN is compiled into LUT artifacts with:
\begin{itemize}
\item $L = 64$
\item Quantization: symmetric int8
\item Interpolation: linear
\item \texttt{boundary\_mode}: closed
\item \texttt{oob\_policy}: clip\_x
\item \texttt{value\_repr}: spline\_component
\item Calibration: 4,096 samples from training set
\end{itemize}

\subsection{Classification Quality Results}

Table~\ref{tab:dos_quality} compares float model and LUT inference on the test set.

\begin{table}[htbp]
\caption{DoS detection: Float vs LUT classification metrics (test set, $n = 69{,}523$)}\label{tab:dos_quality}
\centering
\begin{tabular}{@{}lcccc@{}}
\toprule
Method & Accuracy & Precision & Recall & F1 \\
\midrule
Float (PyTorch) & 0.9899 & 0.9844 & 0.9957 & 0.9900 \\
LUT (NumPy)     & 0.9898 & 0.9840 & 0.9957 & 0.9898 \\
LUT (Numba)     & 0.9898 & 0.9840 & 0.9957 & 0.9898 \\
\bottomrule
\end{tabular}
\end{table}

\textbf{Analysis}: The LUT approximation preserves classification quality with negligible degradation:
\begin{itemize}
\item Accuracy: $-0.0001$ (from 0.9899 to 0.9898)
\item F1: $-0.0002$ (from 0.9900 to 0.9898)
\item Recall: unchanged at 0.9957
\item Precision: $-0.0004$ (from 0.9844 to 0.9840)
\end{itemize}

The NumPy and Numba LUT backends produce identical predictions, confirming consistent implementation.

\begin{figure}[htbp]
\centering
\includegraphics[width=0.75\textwidth]{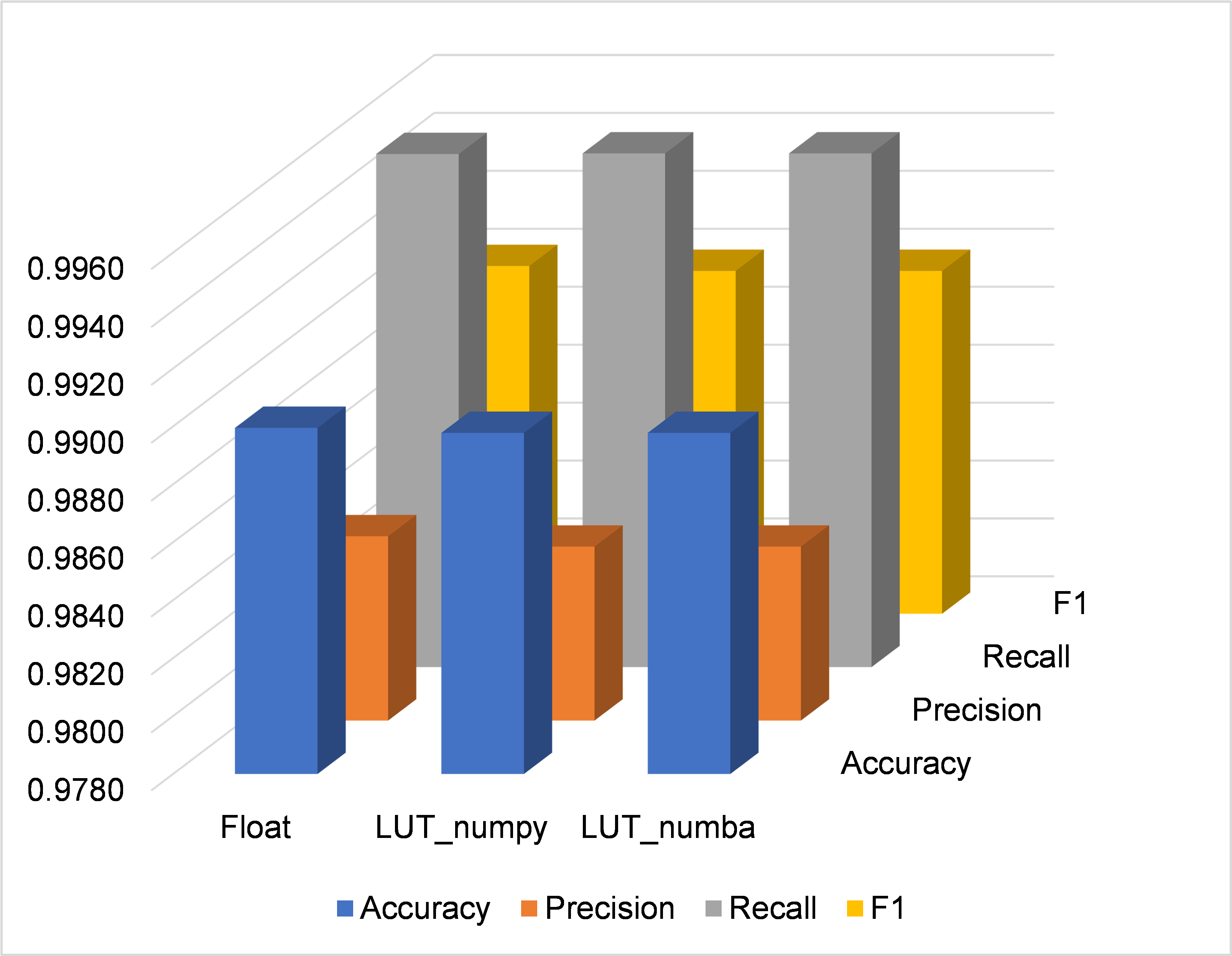}
\caption{Case study (DoS detection): end-to-end classification metrics (Accuracy, Precision, Recall, F1) for the float model and its LUT-compiled variants (NumPy and Numba backends). The differences are imperceptible at this scale.}\label{fig:dos_metrics}
\end{figure}

Figure~\ref{fig:dos_metrics} visualizes the classification metrics.

\subsection{Inference Latency Results}

Table~\ref{tab:dos_speed} reports inference latency for different backends. We distinguish between steady-state inference (artifacts preloaded) and stack-level comparisons (PyTorch vs NumPy/Numba).

\begin{table}[htbp]
\caption{DoS detection: Inference latency (batch = 256, 200 iterations, steady-state)}\label{tab:dos_speed}
\centering
\begin{tabular}{@{}lcccc@{}}
\toprule
Backend & ms/iter & ms/sample & Speedup vs PyTorch$^*$ & Speedup vs B-spline$^\dagger$ \\
\midrule
Float PyTorch      & $226.55 \pm 14.93$ & $0.8850$ & 1.0$\times$ (baseline) & --- \\
LUT NumPy          & $15.26 \pm 1.01$   & $0.0596$ & $14.9\times$ & $\sim$12$\times$ \\
LUT Numba          & $3.52 \pm 0.23$    & $0.0138$ & $64.4\times$ & $\sim$10$\times$ \\
\bottomrule
\end{tabular}

\footnotesize{$^*$Stack-level comparison: combines representation change and backend change.}

\footnotesize{$^\dagger$Honest baseline: LUT vs B-spline in the same backend (from Section~\ref{sec:results}).}
\end{table}

\textbf{Analysis}: Two types of speedup are reported:

\textbf{Stack-level speedups} (14.9$\times$ for NumPy, 64.4$\times$ for Numba) compare PyTorch float to NumPy/Numba LUT. These are valid end-to-end deployment metrics, but they combine representation change with backend change.

\textbf{Honest baseline speedups} ($\sim$12$\times$ for NumPy, $\sim$10$\times$ for Numba) compare LUT to B-spline evaluation in the same backend. This isolates the representation effect and is the more conservative claim.

\textbf{Why LUT is faster}: The LUT kernel replaces spline basis computation (recursive, multiple memory accesses per basis function) with simple table indexing and linear interpolation. This reduces arithmetic complexity and improves cache locality. NumPy benefits because the LUT operations vectorize efficiently; Numba benefits because the tight loop compiles to efficient machine code with fixed data types.

\subsection{Memory Footprint}

Table~\ref{tab:dos_memory} reports memory usage.

\begin{table}[htbp]
\caption{DoS detection: Memory footprint breakdown}\label{tab:dos_memory}
\centering
\begin{tabular}{@{}lrr@{}}
\toprule
Component & Bytes & Fraction \\
\midrule
Float model (PyTorch state dict) & 200,368 & --- \\
\midrule
LUT Layer 0 (78$\times$32 edges) & 1,867,056 & 82.5\% \\
LUT Layer 1 (32$\times$16 edges) & 383,024 & 16.9\% \\
LUT Layer 2 (16$\times$1 edges)  & 12,016 & 0.5\% \\
\midrule
LUT Total & 2,262,096 & 100\% \\
\midrule
LUT / Float ratio & \multicolumn{2}{c}{$11.29\times$} \\
\bottomrule
\end{tabular}
\end{table}

\textbf{Analysis}: The memory overhead is $11.29\times$, which is close to the $10.4\times$ predicted by controlled sweeps for $L = 64$. The first layer dominates (82.5\%) because it has the most edges (2,496 out of 3,024 total).

This has practical implications for model design: the first layer's memory cost scales as $d_{\text{in}} \times d_{\text{hidden}} \times K \times L$. For KAN models with high input dimension, strategies such as feature selection, dimensionality reduction, or edge pruning before LUT compilation can significantly reduce memory.

For memory-constrained deployments, options include:
\begin{enumerate}
\item Reduce $L$ (e.g., $L = 32$ gives $\sim$5$\times$ overhead with slightly higher error)
\item Use float16 for dequantization parameters
\item Prune low-importance edges before LUT compilation
\end{enumerate}

\section{Discussion}\label{sec:discussion}

\subsection{Trade-off Summary}

LUT-KAN provides a simple deployment format for KAN inference with the following trade-offs:
\begin{itemize}
\item \textbf{Speed}: 10--14$\times$ speedup under NumPy, 9.5--11$\times$ under Numba (honest baseline)
\item \textbf{Accuracy}: MAE $\sim 10^{-4}$ at $L = 64$, negligible impact on downstream classification
\item \textbf{Memory}: $\sim$10$\times$ overhead at $L = 64$, dominated by the quantized table
\item \textbf{Complexity}: OOB semantics must be treated as part of the model contract
\end{itemize}

\subsection{When to Use LUT-KAN}

LUT-KAN is most beneficial when:
\begin{enumerate}
\item CPU inference latency is critical (edge devices, real-time pipelines)
\item The KAN model is trained and fixed (deployment phase)
\item Memory overhead of $5$--$20\times$ is acceptable
\item OOB behavior can be specified and validated in advance
\end{enumerate}

LUT-KAN is less suitable when:
\begin{enumerate}
\item Memory is extremely constrained (use lower $L$ or consider pruning)
\item The model will be fine-tuned frequently (LUT must be recompiled)
\item OOB behavior is unpredictable or highly variable
\end{enumerate}

\subsection{Practical Recommendations}

Based on our results, we recommend:

\textbf{Resolution}: Start with $L = 64$ as a balanced point. Use $L = 32$ if memory is tight. Use $L = 128$ only if accuracy requirements are very stringent.

\textbf{Quantization}: Symmetric int8 and asymmetric uint8 perform similarly for the tested layers and calibration distribution. Choose based on implementation convenience.

\textbf{Value representation}: Prefer \texttt{spline\_component} to keep the base branch analytic.

\textbf{Boundary mode}: Use \texttt{closed} when preprocessing clips inputs to the knot domain. Use \texttt{half\_open} for stricter mathematical semantics.

\textbf{OOB policy}: Use \texttt{clip\_x} when OOB events are rare and saturation is acceptable. Use \texttt{zero\_spline} when saturated extrapolation is dangerous.

\textbf{Deployment}: Ensure LUT artifacts are loaded and cached once, not per-inference. The speedup advantage depends on amortizing the artifact loading cost.

\subsection{Deployment Pitfall: Cold-Start Overhead}

In our case study, we also measured ``cold-start'' latency where LUT artifacts are loaded from disk in every iteration. This is an anti-pattern but illustrates the importance of proper deployment:
\begin{itemize}
\item Cold-start NumPy: $175.32 \pm 11.56$ ms/iter (only $1.29\times$ faster than PyTorch)
\item Cold-start Numba: $168.45 \pm 11.11$ ms/iter (only $1.35\times$ faster than PyTorch)
\end{itemize}

The LUT advantage is realized only when artifacts are preloaded and reused for many inferences. In production, this means loading artifacts once at service startup, not per-request.

\subsection{Limitations}

\textbf{Memory overhead for dense layers}: For dense KAN layers with many edges, per-edge tables are large. In the case study, the first layer (78$\times$32 = 2,496 edges) accounts for 82.5\% of total LUT size. Sparse or factored KAN architectures could reduce this.

\textbf{Not a fully integer pipeline}: We quantize LUT values to int8/uint8, but the full layer still uses float32 accumulation and float32 base-branch evaluation. A fully integer pipeline would require additional engineering.

\textbf{Calibration sensitivity}: LUT accuracy depends on the calibration distribution. If the calibration data does not cover the deployment distribution, quantization ranges may be suboptimal.

\textbf{Static compilation}: The LUT is compiled from a fixed trained model. Any model update requires recompilation. Online adaptation is not supported.

\textbf{Single-precision dequantization}: We use float32 for scale and y\_min. Using float16 could reduce memory but may introduce numerical issues for extreme value ranges.

\subsection{Comparison with Alternative Approaches}

Alternative KAN acceleration strategies exist but have different trade-offs:

\textbf{Radial basis functions (RBF)}: Some KAN variants use RBF instead of B-splines~\cite{Abueidda2025}. RBF evaluation can be faster but may sacrifice the local support property of B-splines.

\textbf{Chebyshev polynomials}: Chebyshev-based KAN~\cite{Dong2026,Kumar2026} can use FFT-based evaluation. This is efficient for high-degree polynomials but adds complexity.

\textbf{Network pruning}: Removing low-importance edges reduces both computation and memory. This is orthogonal to LUT compilation and can be combined.

\textbf{Knowledge distillation}: Training a smaller MLP to mimic a KAN could provide faster inference while losing interpretability.

LUT-KAN is complementary to these approaches. It provides a direct compilation path that preserves the exact edge function semantics, which is valuable when interpretability matters.

\section{Conclusion}\label{sec:conclusion}

KAN inference is often limited by spline evaluation cost on CPU. LUT-KAN provides a simple compilation path from trained PyKAN-style KAN layers to segment-wise LUT artifacts with explicit quantization and OOB semantics.

Our controlled experiments show that:
\begin{enumerate}
\item Approximation error follows the expected $O(1/L)$ trend, with MAE $\sim 10^{-4}$ at $L = 64$.
\item Honest baseline speedups are 10--14$\times$ (NumPy) and 9.5--11$\times$ (Numba), confirming genuine representation gains.
\item Memory overhead is approximately $10\times$ at $L = 64$, with the quantized table dominating.
\item OOB behavior depends systematically on boundary mode and OOB policy, making explicit semantics essential.
\end{enumerate}

The DoS detection case study demonstrates that LUT compilation preserves downstream classification quality (F1 drop $< 0.0002$) while reducing steady-state inference latency. The honest baseline speedup is $\sim$10--12$\times$; the stack-level speedup (PyTorch to Numba) reaches $\sim$64$\times$.

The main limitations are increased artifact size for dense layers and sensitivity to OOB behavior. Future work could explore sparse LUT representations, fully integer pipelines, and adaptive quantization based on edge importance.

\backmatter

\bmhead{Acknowledgements}
The author thanks the colleagues who provided feedback on early drafts of this work.

\section*{Declarations}

\textbf{Funding}: This research received no external funding.

\textbf{Conflict of interest}: The author declares no competing financial or non-financial interests.

\textbf{Ethics approval}: Not applicable.

\textbf{Data availability}: The controlled evaluation data is generated programmatically from fixed random seeds. The DoS detection case study uses the CICIDS2017 dataset, available from the Canadian Institute for Cybersecurity (\url{https://www.unb.ca/cic/datasets/ids-2017.html}).

\textbf{Code availability}: All source code is publicly available with fixed release tags for reproducibility:
\begin{itemize}
\item LUT-KAN framework: \url{https://github.com/KuznetsovKarazin/lut-kan} (tag: v1.0.0)
\item DoS case study: \url{https://github.com/KuznetsovKarazin/kan-dos-detection} (tag: v1.0.0)
\end{itemize}

To reproduce Tables~\ref{tab:accuracy}--\ref{tab:sym_asym}:
\begin{verbatim}
git clone --branch v1.0.0 \
  https://github.com/KuznetsovKarazin/lut-kan.git
cd lut-kan && pip install -e .
python scripts/run_experiment.py configs/exp_pykan_lut.yaml
python scripts/generate_experiment_grid.py
# generated configs are saved in configs/generated/
python scripts/run_experiment.py configs/generated/inrange_closed_L16_int8sym.yaml
python scripts/run_experiment.py configs/generated/inrange_closed_L32_int8sym.yaml
python scripts/run_experiment.py configs/generated/inrange_closed_L64_int8sym.yaml
python scripts/run_experiment.py configs/generated/inrange_closed_L128_int8sym.yaml
python scripts/run_experiment.py configs/generated/inrange_closed_L16_uint8asym.yaml
python scripts/run_experiment.py configs/generated/inrange_closed_L32_uint8asym.yaml
python scripts/run_experiment.py configs/generated/inrange_closed_L64_uint8asym.yaml
python scripts/run_experiment.py configs/generated/inrange_closed_L128_uint8asym.yaml
python scripts/collect_results.py --root outputs/exp_runs --outdir outputs/tables
\end{verbatim}

To reproduce Tables~\ref{tab:dos_quality}--\ref{tab:dos_memory}:
\begin{verbatim}
git clone --branch v1.0.0 \
  https://github.com/KuznetsovKarazin/kan-dos-detection.git
python -m pip install -r requirements.txt
python -m pip install numba

python src/train.py
# use the printed run directory under experiment_data/runs/<RUN_ID> as RUN_DIR

python -m src.lut_v2.build_lut ^
  --run-dir RUN_DIR ^
  --out RUN_DIR/lut/L64_sym_int8 ^
  --L 64 ^
  --value-repr spline_component ^
  --scheme symmetric ^
  --dtype int8 ^
  --interp linear ^
  --boundary-mode closed ^
  --oob-policy clip_x ^
  --calib-split train ^
  --num-samples 4096 ^
  --device cpu

python -m src.lut_v2.evaluate_lut ^
  --run-dir RUN_DIR ^
  --lut-dir RUN_DIR/lut/L64_sym_int8 ^
  --backend numba ^
  --threads 1
\end{verbatim}

\textbf{Author contribution}: Oleksandr Kuznetsov: Conceptualization, Methodology, Software, Validation, Formal analysis, Investigation, Data curation, Writing -- original draft, Writing -- review \& editing, Visualization.


\end{document}